\documentclass[11pt]{article}

\usepackage[T1]{fontenc}
\usepackage[utf8]{inputenc}
\usepackage{lmodern}
\usepackage[protrusion=true,expansion=true]{microtype}
\usepackage[margin=1in]{geometry}
\usepackage{amsmath,amssymb}
\usepackage{booktabs,longtable,tabularx,array,calc}
\usepackage{ragged2e}
\usepackage{enumitem}
\usepackage[round,authoryear]{natbib}
\usepackage{xurl}
\usepackage[hidelinks]{hyperref}

\hypersetup{
  pdftitle={WORTH Benchmark: Evidence Discovery under Not-Fully-Cooperative Information -- A Two-Episode Pilot},
  pdfauthor={Bo Zou and Chao Xu},
  pdfsubject={Large language model reasoning evaluation},
  pdfkeywords={large language models, evidence discovery, mystery narratives, reasoning evaluation}
}

\setlist{nosep,leftmargin=*}
\newcolumntype{Y}{>{\RaggedRight\arraybackslash}X}

\DeclareUnicodeCharacter{03B1}{\ensuremath{\alpha}}
\DeclareUnicodeCharacter{00D7}{\ensuremath{\times}}
\DeclareUnicodeCharacter{2212}{\ensuremath{-}}
\DeclareUnicodeCharacter{2260}{\ensuremath{\neq}}
\providecommand{\tightlist}{%
  \setlength{\itemsep}{0pt}\setlength{\parskip}{0pt}}

\begin{document}

\begin{center}
  {\LARGE\bfseries WORTH Benchmark: Evidence Discovery under Not-Fully-Cooperative Information\par}
  \vspace{0.45em}
  {\Large A Two-Episode Pilot\par}
  \vspace{0.8em}
  {\large Bo Zou \qquad Chao Xu\par}
  \vspace{0.4em}
  {\normalsize Bread Studio\par}
  {\normalsize \texttt{bo@BreadMedia.com} \quad \url{https://huggingface.co/BreadStudio}\par}
\end{center}

\begin{abstract}
Existing reasoning evaluations typically provide task-relevant information to models as available premises, rarely examining concurrently how models discover evidence autonomously and how information presentation within narratives guides or obscures judgment. This paper proposes the WORTH Benchmark: models read original mystery narratives without clue annotations, autonomously select key clue sentences, identify case participants, and explain their reasoning process. The narrative materials are designed by professional authors such that evidence facilitating reasoning coexists with narrative arrangements that delay the revelation of truth. We report a pilot experiment under full-story conditions across two episodes, conducting a total of 70 formal runs on models, of which 67 runs are content-scorable. Scoring separately records hits on author-designated key clue sentences and participant selections; natural language explanations are used for case diagnosis and are not scored independently. The results show that clue hits and participant judgments are not synchronized; across 37 content-scorable runs in Episode 02, 31 runs simultaneously selected the same set of incorrect participants, while adjacent suspicion-oriented and exoneration-oriented sentences were selected 25 times and 0 times, respectively. These phenomena illustrate the diagnostic value of component-wise analysis and suggest that not-fully-cooperative informational relationships may influence evidence selection and judgment. This study only implements full-story conditions for two episodes, and does not establish stable model rankings or infer general mechanisms on this basis; human player records have not yet been cleaned, and this version does not report human-model comparisons.
\end{abstract}

\section{1. Introduction}\label{introduction}

There are many varieties of existing large language model reasoning
evaluations, among which premise-provided tasks typically share a common
property: prompt information is helpful. Regardless of how difficult the
problem is, how long the text is, or how complex the reasoning chain is,
the input materials are designed as information to help the model solve
the problem. What the model needs to resolve is how to understand and
invoke this information, without questioning the truthfulness of the
information itself.

Real-world information is rarely presented so helpfully or
transparently. A statement can be entirely factual at the level of
truth, yet mislead the receiver's judgment through selective expression,
omission, and emphasis. This paper uses ``cooperative'' and
``adversarial'' to roughly distinguish between two types of
informational relationships. The adversariality referred to here does
not mean hostile tone, aggressiveness, or false information; rather, it
refers to the directional effect of the text on the reader's judgment in
terms of informational function. This distinction relates to the
Cooperative Principle in pragmatics \citep{grice1975logic}, the theory
of epistemic vigilance \citep{sperber2010epistemic}, and the survey on
the application of Gricean maxims in natural language processing
\citep{krause-vossen-2024-gricean}; its precise positioning within large
language model evaluation will be discussed in the related work section.

This distinction points to two related but distinct issues in current
reasoning evaluations. First, most tasks measure reasoning ability after
evidence has already been provided; ``what is worth becoming evidence''
falls outside the scope of evaluation. Second, lengthening the text or
adding distracting information is not equivalent to taking
not-fully-cooperative informational relationships as an evaluation
condition---test designers may have increased difficulty, but they do
not necessarily examine how information guides judgment through its
presentation. Under the former condition, reasoning can only start from
pre-compiled premises; under the latter condition, evaluation focuses
more on the difficulty of information retrieval rather than how
information presentation affects evidence judgment.

The evaluation proposed in this paper---WORTH---alters both conditions
simultaneously. First, models must autonomously determine which
sentences possess inferential value from unmarked natural narratives;
``finding the starting point of reasoning'' itself constitutes the
evaluation target. Second, what the model confronts is a
not-fully-cooperative information environment---stories have character
trajectories, event sequences, and reasoning cores designed by
professional authors, with these cores subsequently concealed through
narrative techniques. The model must not only autonomously filter
inferential information, but also make judgments in an environment where
cooperative and adversarial information coexist.

Hence, this paper poses an antecedent question: \textbf{When evidence is
not pointed out and the informational environment is not entirely on the
model's side, can the model find the starting point of reasoning?}

Testing this does not require inventing artificial distractions out of
thin air. Mature mystery narratives are inherently texts where
cooperative and adversarial information coexist: fiction authors must
provide sufficient information to deduce the truth, while delaying its
revelation through narrative focus, salience control, and alternative
explanations. The coexistence of cooperative and adversarial information
is not post-hoc noise, but the intrinsic structure of mystery
narratives. Therefore, WORTH adopts professionally written mystery
narratives as the evaluation vehicle. It should be clarified that WORTH
does not claim that stories are equivalent to the real world; rather, it
treats them as a controlled natural narrative experimental
environment---closer to real-world informational relationships than
artificially arranged premises and distractors, while retaining definite
event ground truth and reproducible scoring conditions.

A complete reasoning process encompasses clue localization, clue
interpretation, evidence combination, and final conclusion, but the
current evaluation directly quantifies two of these outputs: after
reading the complete story, the model selects a limited quota of key
clue sentences and determines case participants from candidate
characters. Key clue sentences are the reasoning cores preset during the
story planning phase, which are wrapped in narrative techniques in the
final text with intentionally reduced salience. The evaluation also
requires models to provide natural language explanations of their
reasoning, but currently does not involve independent scoring of
explanation quality.

This paper reports a two-episode pilot experiment. Episode 01 contains
120 sentences (12 key clue sentences), run 5 times each on 6 models;
Episode 02 contains 134 sentences (20 key clue sentences), run 5 times
each on 8 models; across both episodes, there are 70 formal runs in
total. Both episodes have also been deployed as interactive narrative
mystery games, and each has accumulated thousands of human-player score
records; however, human data have not yet completed export and cleaning,
and this version does not report human scores or human-model comparisons
(see §7.3).

The two-episode pilot preliminarily exhibits three classes of phenomena
that are not easily observable in premise-provided evaluations. First,
the performance of the same model across different stages is
inconsistent: a model can identify more key clue sentences yet still
maintain an incorrect theory in participant judgment; conversely, it can
be conservative in participant selection and make fewer errors, yet miss
key evidence. Second, a shared error tendency emerges across different
models---a substantial number of runs across multiple models converge on
the exact same combination of characters, a combination that
superficially enjoys ample evidentiary support but diverges from the
ground truth. Third, models are more likely to select high-salience
suspicion information and more likely to overlook low-salience
counter-evidence and exonerating information. This paper reports these
phenomena as observable behavioral structures, without inferring
internal cognitive mechanisms of the models.

The contributions of this paper span three levels: conceptually,
introducing evidence discovery and not-fully-cooperative informational
relationships into large language model reasoning evaluation;
methodologically, providing an executable evaluation framework based on
professionally crafted natural narratives; empirically, demonstrating
the three aforementioned classes of diagnostic phenomena through a
two-episode pilot experiment.

The current work implements only the full-story condition and should be
understood as a formalized pilot experiment rather than a mature
leaderboard. The goal of this paper is not to prove a universal model
cognitive mechanism using two stories, but to propose and preliminarily
demonstrate a new diagnostic perspective: models need not only reason
upon evidence, but also judge where their reasoning should begin when
evidence is not pointed out and informational relationships are not
purely cooperative.

\section{2. Related Work}\label{related-work}

WORTH's two concurrent conditions---evidence discovery and
not-fully-cooperative informational relationships---are not entirely
groundbreaking research, but align with several academic threads. This
section first clarifies two premises commonly adopted by
premise-provided evaluations, then explains attempts from different
academic traditions to partially alter these premises, and finally
situates WORTH's specific position relative to mystery fiction-related
evaluations.

\subsection{2.1 Common Premises of Existing Reasoning
Evaluations}\label{common-premises-of-existing-reasoning-evaluations}

From GSM8K \citep{cobbe2021training} and MATH \citep{hendrycks2021math}
to GPQA \citep{rein2023gpqa}, FrontierMath
\citep{glazer2024frontiermath}, BIG-Bench Hard
\citep{suzgun2022challenging}, and BIG-Bench Extra Hard
\citep{kazemi-etal-2025-big}, mainstream reasoning benchmarks
differentiate model capabilities by increasing the inferential
complexity of the problem itself---multi-step computation, domain
knowledge thresholds, and combinatorial explosions. These works have
constructed systematic measurement instruments along the dimension of
``whether conclusions can be correctly deduced given the premises,'' but
they typically share two design premises. First, the task objective,
response space, and primary premises required for solution are
pre-organized and provided by the evaluator, without taking open
evidence identification as the core scoring target. Second, the prompt
overall is designed for the purpose of supporting problem-solving; even
when containing irrelevant information or distracting options, it
typically does not take the narrative presentation of factual yet
selective text to guide erroneous judgment as a core experimental
variable. The two conditional shifts proposed in the
Introduction---evidence discovery and not-fully-cooperative
informational relationships---specifically target these two default
premises.

\subsection{2.2 Attempts to Partially Alter These
Premises}\label{attempts-to-partially-alter-these-premises}

\textbf{Long-Context and Narrative Understanding.} NarrativeQA
\citep{kocisky2018narrativeqa} and LiteraryQA
\citep{bonomo-etal-2025-literaryqa} employ books, movie scripts, and
literary works; LongBench v2 \citep{bai-etal-2025-longbench} covers
diverse real-world long documents and long-context tasks. RULER
\citep{hsieh2024ruler} and the Needle-in-a-Haystack series (including
Sequential-NIAH \citep{yu-etal-2025-sequential}; U-NIAH
\citep{gao2025uniah}) measure long-context processing capabilities from
the perspectives of information localization, multi-hop tracking, or
aggregation. While these works increase the difficulty of information
retrieval, the retrieval targets or question-answer pairs are
pre-defined by the evaluators; models do not need to independently judge
``what is worth becoming evidence.'' The relationship between the text
and the model remains cooperative---interference stems from text length
and detail density, rather than from the functional design of the text's
information.

\textbf{Multi-Hop QA and Evidence Detection.} HotpotQA
\citep{yang-etal-2018-hotpotqa} and MuSiQue
\citep{trivedi-etal-2022-musique} require models to combine evidence
across paragraphs, systematically measuring the ability to piece
together reasoning chains from dispersed information. HotpotQA also
requires models to output sentence-level supporting facts and separately
evaluates their accuracy, already incorporating evidence selection into
quantifiable evaluation dimensions. DetectBench
\citep{gu-etal-2024-detectbench} (its early 2023 preprint titled
\emph{Piecing Together Clues}) goes a step further by independently
reporting performance on evidence discovery and final question
answering, presenting cases where ``relevant evidence was found but the
answer was still wrong.'' These works demonstrate that separately
measuring evidence selection and final reasoning is not an entirely
novel attempt. Among them, HotpotQA and MuSiQue primarily use
encyclopedic materials, whereas DetectBench includes detective
narratives and implicit evidence. WORTH likewise poses explicit
questions; its focus lies in incorporating both evidence layout and
narrative obfuscation into material design within original professional
narratives, and recording models' clue selection and participant
judgment.

\textbf{Abductive Reasoning.} αNLI \citep{bhagavatula2020abductive}
formalizes abductive reasoning as the task of selecting the more
plausible explanation between two candidate hypotheses. True Detective
\citep{del-fishel-2023-true} uses 191 long detective mysteries
(averaging \textasciitilde1,200 words) to test models' ability to infer
implicit causal relations; its difficulty is substantially higher than
αNLI, and models indeed need to process information across longer
narratives. However, αNLI's hypothesis space is a predefined binary
choice; True Detective answers specific questions given a puzzle, and
the evaluation output is the final answer, without requiring the model
to explicitly select the evidentiary sentences underlying its reasoning.
The most defensible distinction between these benchmarks and WORTH lies
in their output and scoring designs, rather than in text length.

\textbf{Evidence Selection and Explanation Faithfulness.} ERASER
\citep{deyoung-etal-2020-eraser} established a framework for evaluating
models' ability to ``provide reasoning evidence,'' requiring models not
only to produce answers but also to extract text rationales supporting
those answers (sourced from encyclopedias, medical literature, news,
etc.). Some pipelines in ERASER designed an evaluation mode where
evidence is extracted first and predictions are made solely based on the
extracted content; evidence selection in these designs is not purely
post-hoc explanation. Explanation faithfulness research
\citep{lyu-etal-2024-towards} further focuses on whether models'
generated explanations truly reflect their decision-making process.
ERASER focuses on cross-task rationale extraction and explanation
evaluation; WORTH focuses on original professional narratives, observing
evidence discovery under conditions of author-designed evidence layout
and information obfuscation.

\textbf{Misdirection, Deception, and Adversarial Examples.} Adversarial
NLI \citep{nie-etal-2020-adversarial} increases the difficulty of
natural language inference through human-and-model-in-the-loop
adversarial annotation; GSM-IC \citep{shi2023distracted} inserts
irrelevant sentences into math problems to measure the degree of model
distraction; RECALL \citep{liu2023recall} tests model robustness against
external information that contradicts pre-trained knowledge; LoFa
\citep{shen-etal-2026-truth} tests models' ability to resist fallacious
persuasion; Contrast Sets \citep{gardner-etal-2020-contrast} test
models' decision boundaries through meaningful local perturbations
(typically flipping labels). The commonality among these works is the
artificial construction or rewriting of inputs to probe model
boundaries, of which only some explicitly label such constructions as
adversarial. Such externally applied test conditions differ from the
narrative-intrinsic informational relationships examined by WORTH: the
adversariality in the latter arises from the narrative design of
professional authors and is an inherent structure of mystery narratives
rather than a post-hoc perturbation.

\textbf{The Cooperative Principle and Epistemic Vigilance in
Pragmatics.} The distinction between cooperative and adversarial
relationships mentioned in the Introduction is theoretically aligned
with Grice's \citeyearpar{grice1975logic} Cooperative Principle and the
epistemic vigilance theory of Sperber et al.
\citeyearpar{sperber2010epistemic}. In recent years, researchers have
begun introducing these pragmatic concepts into large language model
research: Krause and Vossen \citeyearpar{krause-vossen-2024-gricean}
surveyed the applications of Gricean maxims in natural language
processing; related studies have examined single-agent trust and
epistemic vigilance regarding external skills \citep{unlu2026trust},
model compliance with versus challenging of users' harmful beliefs
\citep{cheng2026accommodation}, and pragmatic performance as speakers
and listeners in collaboration under asymmetric information
\citep{vanderhoeven2026flout}. These studies touch upon model behavior
both as information producers and receivers, but have been conducted
primarily in conversational or short-text scenarios, without measuring
model performance when confronting information obfuscation in long-form
natural narratives.

\subsection{2.3 Evaluations Related to Mystery Fiction and Detective
Narratives}\label{evaluations-related-to-mystery-fiction-and-detective-narratives}

While the above lines of work touch upon individual aspects of WORTH,
the most directly adjacent work involves existing evaluations centered
on mystery fiction and detective narratives. In recent years, several
studies have introduced mystery novels and detective stories into large
language model evaluation.

DetectiveQA \citep{xu2024detectiveqa} utilizes detective novels
averaging over 100K tokens to construct 1,200 Chinese and English
human-crafted questions and reference reasoning steps, evaluating
long-context reasoning with step-by-step metrics. That study observed
difficulties in evidence retrieval, but the tasks remain triggered by
pre-written questions and do not require models to output a
quota-restricted set of key evidentiary sentences from the full text.

WHODUNIT \citep{gupta2025whodunit} uses full texts of public-domain
mystery novels, requiring models to identify the culprit after reading.
This is one of the works most directly adjacent to WORTH in the task
format of ``reading a relatively complete mystery narrative and
identifying the perpetrator.'' However, two key differences exist:
first, WHODUNIT does not require models to explicitly select key clue
sentences---the ``evidence discovery'' process is absent from the
observable evaluation output, preventing the benchmark from discerning
which evidence path the model took to reach the answer; second, using
public-domain classic mystery novels introduces the risk of data
contamination, as models may have encountered the plots and endings of
these works during training.

WhodunitBench \citep{xie2024whodunitbench} tests model reasoning
capabilities in multi-player interactive murder mystery game settings,
where information is disclosed incrementally through multi-round
dialogue and models must actively ask questions and reason. Its
evaluation conditions differ fundamentally from WORTH's single full-text
reading with structured output. TurnaboutLLM
\citep{yuan-etal-2025-turnaboutllm} extracts long-narrative
contradiction identification tasks from \emph{Ace Attorney} and
\emph{Danganronpa}; candidate testimonies and pieces of evidence are
already explicitly segmented and labeled, requiring models to identify
contradictory pairs across a large combinatorial space rather than
delineating evidentiary sentences from unmarked narratives on their own.
The computational approach proposed by Wagner et al.
\citeyearpar{wagner2025fairplay} formalizes surprise, coherence, and
``fair play'' in narratives using an information-theoretic framework;
its focus is on evaluating whether a work plays fair with readers, not
on measuring whether readers or models can reason successfully under
fair yet cognitively obscured conditions.

In summary, existing evaluations related to mystery fiction have
introduced this genre into large language model testing, but their
evaluation outputs consist primarily of final reasoning conclusions
(identifying the culprit, answering questions), without requiring models
to explicitly select the evidentiary sentences underlying their
reasoning and incorporating them into itemized scoring. Furthermore,
materials used in these works are mostly existing works, lacking
experimental control where authors actively design information
obfuscation conditions from the creative inception stage.

\subsection{2.4 Positioning of WORTH}\label{positioning-of-worth}

Synthesizing the above research trajectories, separately measuring
evidence selection and reasoning conclusions is not an entirely novel
attempt (as discussed in §2.2 regarding HotpotQA, DetectBench, ERASER,
etc.), and multiple explorations have employed mystery fiction as an
evaluation vehicle. WORTH's positioning lies in concurrently realizing
the following three conditions:

\begin{enumerate}
\def\labelenumi{\arabic{enumi}.}
\tightlist
\item
  \textbf{Autonomous Evidence Discovery}---Models independently select
  clue sentences from unmarked, complete narratives; the selections
  themselves are incorporated into scoring and recorded separately from
  final participant judgments;
\item
  \textbf{Not-Fully-Cooperative Informational Environment}---Materials
  are designed from scratch by professional authors; the coexistence of
  cooperative and adversarial information constitutes the narrative's
  intrinsic structure rather than an externally appended disturbance or
  noise;
\item
  \textbf{Itemized Diagnostic Output}---Clue hits and participant
  attributions are scored separately, allowing desynchronization between
  ``what was found'' and ``what was concluded based on what was found''
  to be directly observed.
\end{enumerate}

To the authors' knowledge, no existing benchmark satisfies all three
conditions simultaneously. Each condition appears individually or
partially in existing literature: DetectBench separately reports
implicit evidence discovery and task answering, while WORTH further
emphasizes the joint design of evidence layout and obfuscation in
original professional narratives; WHODUNIT uses full-length mystery
novels but does not require explicit clue sentence selection; True
Detective uses long-form detective puzzles, but its evaluation output is
the final answer. WORTH's contribution lies in combining these three
conditions into an executable evaluation framework, rather than claiming
to pioneer any single concept in isolation.

It should be noted that WORTH does not claim to have exhaustively
covered the literature in any of the aforementioned research domains.

\section{3. The WORTH Evaluation
Framework}\label{the-worth-evaluation-framework}

The Introduction raised two conditional shifts: autonomous evidence
discovery and not-fully-cooperative informational relationships. This
section elaborates on how these two conditions are translated into an
actionable and analyzable evaluation design, and explains why
professionally authored mystery narratives serve as an appropriate
experimental vehicle.

\subsection{3.1 Autonomous Evidence Discovery and the Target Capability
Chain}\label{autonomous-evidence-discovery-and-the-target-capability-chain}

WORTH requires models to complete three separable stages. The first
stage is clue localization: after reading a complete natural narrative,
the model independently assesses which sentences among all sentences
possess inferential value for identifying case participants, and selects
a limited quota of sentences as its evidentiary basis. The second stage
is analysis and reasoning: the model conducts analysis based on the
selected clue sentences---analysis here is understood broadly,
potentially involving logical deduction, physical commonsense,
sociological knowledge, or combinations of multiple reasoning
modalities. The third stage is participant attribution: based on the
analytical results, the model identifies which candidate characters
appearing in the story are case participants.

Current scoring directly quantifies the first and third stages. The
score for clue sentence selection is based on the overlap between
sentences selected by the model and author-designated key clue sentences
(overlap with the author-designated key-sentence set K). Clue sentence
selection employs a quota mechanism: selecting non-key sentences incurs
no direct point deduction, but consumes the limited quota---the penalty
is embedded in opportunity cost rather than explicit deduction. The
participant attribution stage employs an explicit reward-and-penalty
scheme: correctly identifying case participants yields rewards, while
erroneously selecting non-participants incurs penalty deductions. The
two episodes differ in the specific weight designs for clues and
participants (frozen weights for each episode are detailed in §4.5), but
share an overarching framework principle: clue sentences represent
intermediate steps whose value is ultimately manifested in participant
attribution; participant attribution is the terminal judgment, where an
erroneous conclusion represents a definitive failure of judgment.

The second stage---analysis and reasoning---is currently not
incorporated into quantitative scoring. The evaluation requires models
to provide reasoning analysis in natural language, but does not
independently score the analysis itself. This decision stems not only
from operational considerations (reliable scoring of natural language
analysis requires either expensive manual coding or introduces the
circular dependency of another model judge), but also from an assessment
of the function of the analysis stage: analysis is a processual output
whose quality is ultimately reflected in clue selection and participant
attribution outcomes; forcing independent quantification onto it risks
distorting an open reasoning process into optimization for scoring
heuristics. At the present stage, the primary value of the analytical
output lies in providing interpretable process information for
diagnostic evaluation---such as determining whether the model understood
the function of a particular clue sentence, or tracing the reasoning
trajectory behind an erroneous attribution---rather than serving as an
independent scoring dimension. Whether this design choice warrants
revision in future work is left to subsequent empirical accumulation.

\subsection{3.2 Not-Fully-Cooperative Informational
Relationships}\label{not-fully-cooperative-informational-relationships}

The Introduction used ``cooperative'' and ``adversarial'' to roughly
distinguish between two types of informational relationships. This
section further elaborates on this distinction.

Adversariality refers to situations where, although information in the
text is entirely truthful at the factual level, its
presentation---including narrative placement, salience, and causal
framing---functionally impairs the reader's ability to form correct
judgments. Adversariality is not false information, not hostile tone,
and not random noise; rather, it represents informational control
achieved through narrative design by professional authors.

Take an illustrative design from Episode 02. One case participant in the
story appears with a limp in his left leg after a fight and subsequently
walks with a crutch (S081). The story text factually states: ``Geng Baishun
limped in his left leg, and the injury was quite severe; for many days
afterward, I saw him propping a crutch under his left arm, grimacing
with pain at every move.'' One detail is that, despite the injury being
in his left leg, he supports the crutch with his left hand, whereas
under normal circumstances one would use the contralateral hand. This
anomaly is deliberately designed by the author---the injury is feigned,
and the crutch serves another purpose---yet the story text does not inform
the reader of this. What the reader sees is a plausible causal chain:
fight → injury → crutch. Reader attention is captured by the fight event
itself, rendering the crutch a natural consequence of the injury rather
than a clue requiring independent scrutiny. The critical detail (left
leg injured yet crutched on the left hand) is already presented in the
story text, but why it is anomalous and how it relates to the case must be
autonomously discovered and interpreted by the reader.

Cooperative information (evidence sufficient to deduce the truth) and
adversarial information (narrative arrangements reducing the salience of
critical evidence) coexist within the same text. This is not post-hoc
artificial noise, but the intrinsic structure of mystery
narratives---mature mystery narratives inherently require authors to
achieve two tasks simultaneously: provide evidence in principle
sufficient to deduce the truth, and delay the revelation of truth
through narrative techniques.

\subsection{3.3 Why Professionally Crafted Mystery Narratives Are
Appropriate Experimental
Vehicles}\label{why-professionally-crafted-mystery-narratives-are-appropriate-experimental-vehicles}

WORTH uses original, professionally authored mystery narratives designed
from scratch as evaluation materials, rather than existing mystery
novels or artificially assembled distracting texts. This choice is
grounded in three reasons.

First, designing from scratch provides design-level support for the
benchmark's construct validity. WORTH measures whether models can
discover the reasoning cores preset by the author---key clue sentences
are designated as reasoning cores prior to story composition, and the
narrative unfolds around them. If commercially published novels were
selected, pinpointing reasoning clues accurately would be difficult,
because within a finished mystery novel, determining which information
was intentionally planted as critical clues and which emerged as
incidental usable details during narration cannot be ascertained from
the text alone. Designing from scratch means that the inferential
function, salience positioning, and role within the evidence chain of
every key clue sentence are established during the creative phase.

Second, original materials mitigate the exposure risk associated with
directly utilizing existing mystery fiction. Classic works in the public
domain may have entered model training datasets, where models' memory of
plots and endings would contaminate measurement; crafting from scratch
makes material provenance and answer design more traceable, although it
cannot guarantee that models have never encountered the stories prior to
evaluation. The timeline relationship between the public release of
materials and model testing across both episodes is detailed in §4.6.

Third, mystery narratives naturally provide a textual environment where
cooperative and adversarial information coexist. If evaluation materials
simply insert distractor sentences randomly into normal text, the
distraction itself is artificial---this measures noise detection
capability. Informational relationships in mystery narratives are not
noise, but functional components of narration.

Precisely because the informational structure of such materials is the
product of narrative design rather than post-hoc annotation, its
production process differs from problem writing or crowdsourced
annotation in conventional benchmarks. The latter can typically be
completed by a team from a single professional background; in contrast,
evaluations that employ natural narratives as vehicles and concurrently
undertake evidence discovery and the measurement of
not-fully-cooperative informational relationships require the
coordinated integration of three professional competencies typically
residing in different domains throughout material production.

The first is narrative design capability. At their best, evaluation
materials are natural, plausible, and literarily competent mystery
narratives, allowing cooperative and adversarial information to coexist
organically within the same text.

The second is evaluation methodology capability. The narrative structure
must serve quantifiable cognitive discrimination. The quantity, weight
distribution, and salience gradients of key clue sentences are not
literary judgments, but measurement design decisions; distractor
characters require sufficient superficial evidentiary pointing while
remaining excludable by specific low-salience
counter-evidence---appearing strongly suspect yet definitively
exonerable. If a creative team lacks this capability, the story may be
compelling in itself, yet fail to generate measurement signals with
discriminatory power across models.

The third is diagnostic analytical capability. Evaluation is not a
one-off procedure: model performance in pilot experiments---such as
common explanatory attractors across models, or asymmetric processing
between high-salience suspicion information and low-salience
counter-evidence---needs to be mapped back to material design. This
feedback loop is essential to determine whether observed phenomena stem
from the salience design of the story text itself or reflect systematic model characteristics,
and to calibrate the evidentiary structure of subsequent episodes
accordingly. Lacking this capability, the benchmark can neither iterate
and improve from pilot results nor disentangle item effects from model
effects.

In this project, these three capabilities were borne by the same small
team---the narrative author participated in the top-level design of the
evaluation architecture, while the evaluation designer participated in
the narrative's evidence layout and exclusion criteria setup. Such
integration currently lacks standardized workflows to follow. Further
formalization of narrative design methodology---including how to
systematically design salience gradients and how to feed back model
results into evidence structure calibration---constitutes a future
direction for this project and is not expanded upon in this paper's
pilot experiment.

\subsection{3.4 Current Implementation Scope and Subsequent Verification
Conditions}\label{current-implementation-scope-and-subsequent-verification-conditions}

WORTH's design differentiates between three conditions: full-story,
evidence-cueing, and evidence-modification, intended to observe response
variations under different conditions.

The first verification layer is the full-story condition, which is also
the only condition implemented in this paper: models read unmarked,
complete narratives and autonomously complete clue selection, reasoning
analysis, and participant attribution.

The second verification layer is the evidence-cueing condition:
retaining the complete story while explicitly marking the key clue
sentences designated by the author, comparing participant attribution
before and after prompting. This tests whether prompting key evidence
alters judgment, though such a comparison alone cannot attribute
differences entirely to evidence discovery.

The third verification layer is the evidence-modification condition:
modifying key evidence while maintaining narrative coherence and
altering the answer accordingly, observing whether model judgments
change alongside evidentiary shifts. This condition tests model
responsiveness to textual evidence variations.

The latter two verification layers have not yet been implemented and do
not constitute empirical evidence in this paper.

\section{4. Design of the Two-Episode Pilot
Experiment}\label{design-of-the-two-episode-pilot-experiment}

This section describes the experimental design, response protocol,
operational conditions, scoring rules, and data processing criteria of
the two-episode pilot experiment. The two episodes differ in story
content, evidence structure, and scoring parameters; raw scores cannot
be aggregated or directly compared across episodes.

\subsection{4.1 Task Materials}\label{task-materials}

Each experimental episode uses an original Chinese mystery narrative as
task material. Stories are presented to subjects as numbered natural
sentences, with each sentence preceded by a localization identifier
ranging from \texttt{{[}S001{]}} to \texttt{{[}Snnn{]}}; prompt
instructions explicitly clarify that identifiers serve solely for
localization and do not imply whether a sentence constitutes a clue.

Episode 01 comprises 120 natural sentences, revolving around a homicide
occurring at a township grain depot and involving 6 candidate
characters. Episode 02 comprises 134 natural sentences, continuing
another homicide within the same narrative universe and involving 9
candidate characters. The two episodes differ in narrative complexity,
evidence volume, character scale, and scoring weights: Episode 02
features more candidate characters (9 vs.~6) and key clue sentences
increase from 12 to 20. The weight designs of both episodes emphasize
different aspects: EP01 assigns uniform weights to clues (3 points
each), differentiates rewards for correct characters (7/5/5), and
applies a uniform deduction for incorrect characters (-5); EP02
differentiates clue weights (1--4 points), applies uniform rewards for
correct characters (2 points each), and differentiates deductions for
incorrect characters (-2 to -8, penalizing the most misleading
distractor characters more heavily). These differences reflect
evaluation iteration rather than an intentional construction of a
difficulty gradient.

Both stories have been deployed on Bilibili's Toy platform as
interactive reasoning games. Players read the story incrementally
through the game interface and submit a limited quota of clue sentences
and case participant selections. As of the data freeze, each episode had
accumulated thousands of human player score records. Platform cumulative
records have not yet completed export and cleaning, and this version
does not treat them as human-model comparison samples (see §4.8).

The public game pages are
\href{https://www.bilibili.com/toy/zlVdAVXGwnnXEVIM/index.html}{Episode
01} and
\href{https://www.bilibili.com/toy/jUFTfiV4ddkcvfNj/index.html}{Episode
02}. These pages illustrate the player response environment; the current
online game content is not a substitute for the frozen prompts and
scoring versions used in the model experiments.

\subsection{4.2 Design Status of Key Clue
Sentences}\label{design-status-of-key-clue-sentences}

Key clue sentences are inferential hubs preset by the author during the
design of the story world, character trajectories, and event sequences,
rather than post-hoc annotations made after text completion. Each key
clue sentence is documented in the creative archive with its
corresponding original sentence, a brief description of its inferential
function, and its position within the evidence chain.

It should be noted that the author's set of key clue sentences (\(K\))
represents the reasoning cores designated by the author, and does not
represent the sole exhaustive set of all valid evidence in the story. An
unmarked natural narrative may contain sentences unannotated by the
author that nonetheless carry auxiliary inferential value, and multiple
paths may lead to the correct conclusion. Current scoring is based on
the overlap between sentences selected by the model and \(K\) (overlap
with the author-designated key-sentence set K); hence, it measures the
extent to which the model hits the author's preset reasoning cores,
rather than whether it exhausts all possible reasoning pathways.

\subsection{4.3 Response Protocol}\label{response-protocol}

Subjects receive the complete numbered story text alongside a structured
response prompt. The prompt instructs subjects to complete three tasks:
select key clue sentences from all sentences, determine case
participants from candidate characters, and provide reasoning analysis
in natural language.

The output format is a structured object containing seven mandatory
fields: episode identifier, array of clue sentence identifiers, array of
character identifiers, array of sentence-by-sentence clue explanations,
combined evidence analysis, final conclusion, and confidence score. The
instructions include an output self-check checklist, requiring clue
identifiers to be non-duplicate and drawn from the story body, character
identifiers to be non-duplicate and drawn from the candidate list, and
sentence-by-sentence explanations to map one-to-one with clue
identifiers.

Clue selection is subject to an upper bound: Episode 01 permits
selecting at most 12 sentences, and Episode 02 permits at most 20
sentences; subjects may select fewer, but cannot exceed the quota.
Character selection adopts a multi-select mode: subjects may select
none, select one, or select multiple characters, with the prompt not
disclosing the correct number of participants. The instructions for
Episode 02 additionally provide an operational definition of ``homicide
participant''---``including those who planned, executed, or provided
critical assistance to the homicide, not limited to those who directly
carried out fatal acts''---and advise subjects that they need not
exhaust the clue quota if confidence is insufficient.

The nine rules in the instructions also stipulate: utilize only the
information provided in the current prompt; do not access the internet,
search, or use external tools; the narrative perspective is reliable and
narrator deceit need not be considered; do not pose questions to the
test administrator or provide multiple alternative answers; and output
only a single valid structured object without any additional text.

\subsection{4.4 Scoring Information
Disclosure}\label{scoring-information-disclosure}

A deliberate design discrepancy exists between the scoring information
available to subjects during testing and the complete frozen rules
disclosed in the paper.

Information known to subjects includes: the total number of key clue
sentences and the selection ceiling; that selecting correct clues adds
points while selecting incorrect ones incurs no deduction; that
selecting correct characters adds points while selecting incorrect ones
incurs deductions; and that the prompt does not disclose the correct
number of participants. Information unknown to subjects includes: the
differentiated weights of individual key clue sentences (Episode 02),
the differentiated penalty magnitudes for different incorrect
characters, and the complete payoff matrix.

This arrangement represents an explicit measurement design decision. If
subjects knew the complete payoff matrix in advance---for instance,
knowing that misselecting a certain character incurs a severe penalty
while another character incurs a mild penalty, or that certain clue
sentences carry much higher scores than others---response behavior might
shift from narrative reasoning to risk control under a known payoff
matrix. Subjects could avoid costly errors by selecting fewer
characters, or optimize expected scores by prioritizing guesses on
high-value clue sentences, thereby introducing strategic variance
unrelated to narrative reasoning. Concealing differentiated weights and
penalty matrices aims to preserve natural narrative judgment conditions
as much as possible and reduce construct-irrelevant variance. This paper
does not frame this arrangement as having measured ``pure reasoning
ability,'' but rather as having controlled for reasoning-irrelevant
strategic factors within a reasonable scope.

Concealing this information from subjects while disclosing it to paper
readers is not contradictory: the former protects measurement
conditions, while the latter ensures transparency and reproducibility.

\subsection{4.5 Scoring Rules}\label{scoring-rules}

The scoring structures of both episodes adhere to the same logic---clue
hits receive positive points, while participant attribution receives
rewards for correct choices and penalties for incorrect ones---but
differ in specific parameters.

Episode 01 adopts uniform weights. 12 key clue sentences are worth 3
points each; selecting a key sentence earns points, omitting it incurs
no deduction, and selecting a non-key sentence likewise incurs no
deduction but consumes the limited quota. Rewards for the 3 correct
characters are 7 points, 5 points, and 5 points, respectively; incorrect
characters incur a uniform deduction of 5 points. The maximum score is
53 points, and the theoretical minimum is −15 points.

Episode 02 adopts differentiated weights. The 20 key clue sentences
range from 1 to 4 points each, reflecting differing inferential
difficulty and importance within the evidence chain; the clue component
totals 41 points. The 3 correct characters receive 2 points each.
Penalties for incorrect characters vary by character: the most plausible
distractor characters incur the heaviest deductions (up to −8 points),
while characters easier to eliminate incur milder deductions (−2
points). The maximum score is 47 points.

The asymmetry in penalty mechanisms between clue selection and
participant attribution---where selecting non-key sentences carries no
direct point deduction while selecting incorrect characters incurs
explicit penalties---reflects their distinct roles within the reasoning
process. Clue sentences represent intermediate steps whose value is
ultimately realized in participant attribution; participant attribution
is the terminal judgment, where an erroneous conclusion represents a
definitive failure of judgment. The current total score represents a
gameplay-equivalent score, and the clue score represents overlap with
the author-designated key-sentence set K; neither is equivalent to a
comprehensive measurement of complete reasoning capability.

\subsection{4.6 Models and Run
Configurations}\label{models-and-run-configurations}

Episode 01 tested 6 models, and Episode 02 tested 8 models. Model
selection covered major commercial large language models publicly
accessible at the time of data freezing; each group was run according to
product portals and tier configurations recorded at the time, which were
not uniformly set to the ``highest'' tier.

Each model was run 5 times, yielding 70 formal runs across both
episodes. The operational conditions recorded in the experiments were:
conducted in fresh dialogue sessions without inheriting prior
conversation history; web search, external tools, product memory, and
personalization settings disabled; complete prompts (numbered story text
and response instructions) sent in a single turn without multi-turn
interaction. Configuration screenshots and response screenshots exist
for some models, with screenshots providing varying scopes of
substantiation across toggles.

The model lineups across the two episodes partially overlap: Episode 01
includes GPT-5.6 Sol, Claude Opus 4.6, GLM-5.2, DeepSeek-V4-Flash, Kimi
K3, and Gemini 3.1 Pro; Episode 02 includes GPT-5.6 Sol, Claude Opus
4.6, Gemini 3.1 Pro, GLM-5.3, Kimi K3, DeepSeek-V4-Pro, Hy4 preview, and
MiniMax-M3. Certain model versions and access portals shifted between
episodes: for example, GLM-5.2 changed to GLM-5.3, and DeepSeek-V4-Flash
changed to DeepSeek-V4-Pro; Opus shifted from Antigravity's Thinking to
Claude.ai's High + Extended, and Kimi shifted from official website
``Advanced'' to WorkBuddy ``High''. EP02 DeepSeek's tier was not
recorded separately, and MiniMax's tier was recorded as n/a; for EP01
GLM, configuration and execution screenshots showed ``Highest'' and
``High'' respectively, and this paper follows the execution record using
``High''.

EP01 was first released publicly as a game on 2026-08-14 at 17:18
(UTC+8), and its model testing was also conducted on 2026-08-14. EP02 went online on 2026-08-21, and
model testing was conducted from 2026-08-30 to 2026-08-31. Existing
records cannot guarantee that test materials were unreleased prior to
testing, nor can they determine whether model pretraining ingested the
publicly released materials.

\subsection{4.7 Handling of Format
Validity}\label{handling-of-format-validity}

The response prompt requires subjects to output a valid structured
object. In practical processing, whether a response is parseable as JSON
and whether the selections within satisfy identifier and quantity
constraints are recorded separately. That a response is directly
parseable, or that complete JSON can be extracted from surrounding
explanatory text, does not automatically imply that selections are
valid; when overall JSON is unparseable, if valid selection arrays can
be directly read from the raw text, content scoring can still be
computed without repairing the raw response.

Among the 30 runs in Episode 01, 22 were parseable overall, 7 contained
embedded parseable JSON, and 1 had invalid overall JSON but selections
could be directly read; all 30 were content-scorable. Among the 40 runs
in Episode 02, 26 were parseable overall, 9 contained embedded parseable
JSON, and 5 had invalid overall JSON, among which selections in 4 runs
could still be directly read and scored. Out of 40 runs, 3 were
excluded: Gemini response05 and Hy4 response05 had parseable JSON but
contained non-existent identifiers S207 and S146, respectively; Kimi
response02 contained illegal expressions in selection fields and could
not be read as valid identifier arrays. Consequently, EP02 yielded 37
content-scorable runs; excluded runs are not scored as zero.

Across both episodes, responses fully compliant with the output protocol
numbered 22 and 25, respectively. Format compliance and content scores
are reported separately.

\subsection{4.8 Criteria for Human Player
Data}\label{criteria-for-human-player-data}

Both narrative reasoning games have accumulated thousands of human
player score records, but cumulative platform records cannot
automatically be equated with independent, valid human samples. The
human data criteria planned for this paper use cleaned valid
first-attempts: for multiple attempts by the same player identifier
within the same episode, only the initial submission is retained; trial
plays and obvious anomaly records are excluded.

Operational conditions differ between humans and models. Human players
read and submit incrementally through the game interface, whereas models
receive the complete prompt in a single turn; human response times are
uncontrolled, whereas model inference times are determined by respective
platforms; human player counts vastly exceed the 5 runs per model.
Therefore, subsequent human analyses will primarily provide baseline
references for score distributions and error patterns on the same tasks,
examining whether specific phenomena also appear in human responses,
rather than treating large-sample human records and 5 runs per model as
a symmetric statistical comparison. The aforementioned cleaning has not
yet been completed, and this version does not report human data analysis
results (see §7.3).

\subsection{4.9 Semantic Boundaries of ``Case
Participant''}\label{semantic-boundaries-of-case-participant}

The core questions of the two episodes are ``Who brought about the
homicide?'' (Episode 01) and ``Which individuals are homicide
participants?'' (Episode 02). The natural semantic scope of the term
``participant'' may induce divergent interpretations across models and
humans. The prompt for Episode 02 provided an operational
definition---including planners, executors, or key facilitators---yet
this definition still leaves borderline ambiguities. Furthermore, EP02
reviews the death of Uncle Six while also narrating the subsequent
deaths of Qin Guanggui and Wang Yonghong; the correct set of characters
for this episode corresponds to the latter two deaths, but the prompt
only asks ``Which individuals are homicide participants?'' without
explicitly delimiting the scope of the case. Raw responses in some runs
analyzed both Uncle Six's case and subsequent events. This paper treats
this as a construct boundary of the pilot experiment, rather than
attributing all attribution errors related to the definition of
``participant'' to failures of model reasoning. Relevant explanations
should be evaluated in conjunction with the scope of the case and the
meaning of participant actually addressed in the model's response.

\subsection{4.10 Cross-Episode Reporting
Principles}\label{cross-episode-reporting-principles}

Because the two episodes use different stories, different evidentiary
structures, different candidate character sets, and different scoring
parameters, raw scores cannot be directly summed, nor can they be used
to determine which episode is ``more difficult.'' In the Results
section, the two episodes are first reported separately, with
cross-episode comparisons conducted solely on structural phenomena
observable in both episodes (such as process dissociation, common
attractors, and evidence asymmetry). Cross-episode comparisons are based
on observable behavioral patterns and do not rely on numerical
comparisons of raw scores.

\section{5. Results: Behavioral Differences Exhibited in the Pilot
Experiment}\label{results-behavioral-differences-exhibited-in-the-pilot-experiment}

This section reports the results of the two-episode pilot experiment.
Sections 5.1 and 5.2 report intra-episode behavioral structures, and
Section 5.3 extracts structural patterns shared across both episodes.
All numbers are traceable to frozen scoring data. It must be reiterated:
the scale of 70 formal runs across two episodes suffices only to support
preliminary diagnostics, not to establish stable model rankings or infer
generalized cognitive mechanisms.

\subsection{5.1 Episode 01}\label{episode-01}

EP01 collected a total of 30 model runs (6 models × 5 independent runs).
The full text contains 120 sentences, with 12 author-designated key clue
sentences (3 points each, maximum 36 points) and 6 candidate characters
(selecting ``I'' awards 7 points, selecting Father or Director Huang
awards 5 points each, with a 5-point deduction for each misselected
character). The participant component has a maximum of 17 points, and
the gameplay-equivalent total score has a maximum of 53 points.

The table below reports each model's mean gameplay-equivalent total
score and subscale means.

{\def\LTcaptype{none} % do not increment counter
\begin{longtable}[]{@{}lrrr@{}}
\toprule\noalign{}
Model & Total Score Mean & Clue Score Mean & Participant Score Mean \\
\midrule\noalign{}
\endhead
\bottomrule\noalign{}
\endlastfoot
GPT-5.6 Sol & 11.20 & 10.20 & 1.00 \\
Claude Opus 4.6 & 11.60 & 9.60 & 2.00 \\
GLM-5.2 & 15.00 & 9.00 & 6.00 \\
DeepSeek-V4-Flash & 14.60 & 6.60 & 8.00 \\
Kimi K3 & 9.40 & 8.40 & 1.00 \\
Gemini 3.1 Pro & 11.20 & 7.20 & 4.00 \\
\end{longtable}
}

Note: Means for each model are calculated across 5 content-scorable
runs. Total score = clue score + participant score; participant score
includes rewards for correct selections and deductions for incorrect
selections. Maximum scores for total score, clue score, and participant
score are 53, 36, and 17 points, respectively.

\subsubsection{5.1.1 Finding Clues Does Not Imply Identifying
Participants}\label{finding-clues-does-not-imply-identifying-participants}

In EP01, performance in hitting key clue sentences was not synchronized
with participant selection: the model that identified the most clues
yielded the lowest mean for correct participants. This paper uses
\textbf{process dissociation} to describe this desynchronization of
output metrics, without asserting that internal model processes have
been isolated. The table below presents the means of both metrics.

{\def\LTcaptype{none} % do not increment counter
\begin{longtable}[]{@{}
  >{\raggedright\arraybackslash}p{(\linewidth - 10\tabcolsep) * \real{0.1500}}
  >{\raggedright\arraybackslash}p{(\linewidth - 10\tabcolsep) * \real{0.2000}}
  >{\raggedright\arraybackslash}p{(\linewidth - 10\tabcolsep) * \real{0.2000}}
  >{\raggedright\arraybackslash}p{(\linewidth - 10\tabcolsep) * \real{0.1800}}
  >{\raggedright\arraybackslash}p{(\linewidth - 10\tabcolsep) * \real{0.1600}}
  >{\raggedright\arraybackslash}p{(\linewidth - 10\tabcolsep) * \real{0.1100}}@{}}
\toprule\noalign{}
\begin{minipage}[b]{\linewidth}\raggedright
Model
\end{minipage} & \begin{minipage}[b]{\linewidth}\raggedright
Key Clue Hits Mean (/12)
\end{minipage} & \begin{minipage}[b]{\linewidth}\raggedright
Mean No.~of Correct Participants Selected (/3)
\end{minipage} & \begin{minipage}[b]{\linewidth}\raggedright
Mean No.~of Incorrect Participants Selected
\end{minipage} & \begin{minipage}[b]{\linewidth}\raggedright
Five-Run Total Scores
\end{minipage} & \begin{minipage}[b]{\linewidth}\raggedright
Standard Deviation
\end{minipage} \\
\midrule\noalign{}
\endhead
\bottomrule\noalign{}
\endlastfoot
GPT-5.6 Sol & 3.4 & 0.8 & 0.6 & 14, 17, 4, −4, 25 & 11.3 \\
Claude Opus 4.6 & 3.2 & 1.0 & 0.6 & 22, 7, 4, 6, 19 & 8.3 \\
GLM-5.2 & 3.0 & 1.4 & 0.2 & 22, 16, 20, 14, 3 & 7.4 \\
Kimi K3 & 2.8 & 1.0 & 0.8 & 19, 1, 12, 8, 7 & 6.7 \\
Gemini 3.1 Pro & 2.4 & 1.4 & 0.6 & 6, 19, 19, 6, 6 & 7.1 \\
DeepSeek-V4-Flash & 2.2 & 1.6 & 0.0 & 11, 13, 22, 8, 19 & 5.8 \\
\end{longtable}
}

In the table, ``Standard Deviation'' measures the dispersion of scores
across the five runs of the same model (sample standard deviation,
rounded to one decimal place). GPT-5.6 Sol had the largest standard
deviation (11.3), with five total scores ranging from −4 to 25, a span
of 29 points; DeepSeek-V4-Flash had the smallest standard deviation
(5.8), being the most stable among the six models. Because total scores
can be negative (GPT exhibited negative scores in EP01), coefficient of
variation (CV = standard deviation ÷ mean) is not used here, and raw
standard deviations alongside the five individual scores are reported
for direct comparison.

GPT-5.6 Sol identified the most key clue sentences (mean 3.4/12), but
had the lowest mean for correct participants (0.8/3), with a mean of 0.6
incorrect participants per run. DeepSeek-V4-Flash hit the fewest clues
(mean 2.2/12), but had the highest mean for correct participants
(1.6/3), with zero incorrect selections across all five runs. These
itemized outputs exhibit divergent behavioral profiles.

\subsubsection{5.1.2 Case Observations of Erroneous
Judgments}\label{case-observations-of-erroneous-judgments}

The following cases illustrate different error manifestations; mutual
exclusivity classification or frequency counting was not performed
across all responses.

\begin{itemize}
\tightlist
\item
  \textbf{Participant Attribution Drift.} GPT-5.6 Sol's response04 noted
  that Uncle Six did not commit suicide and that Qin Guanggui and Wang
  Yonghong were involved in the grain depot deficit, yet deduced that
  ``Qin Guanggui and Wang Yonghong\ldots{} acted together,'' ultimately
  selecting only these two non-correct characters.
\item
  \textbf{Extratextual Causal Completion.} GLM-5.2's response05
  conjectured that the cupping apparatus ``might have been tampered
  with'' and further speculated that someone used it to administer
  drugs; the story text provided no evidence of apparatus tampering or drug
  administration, and this run also erroneously selected Qin Guanggui.
\item
  \textbf{Omission of Correct Participant.} The first-person narrator
  ``I'' (R04) belongs to the author's participant set, but was selected
  in 0 out of 30 runs. Some responses mentioned ``my'' cupping act, yet
  did not include R04 among participants; this boundary issue is
  discussed in §6.5.
\end{itemize}

\subsubsection{5.1.3 Gemini 3.1 Pro: Scores at Only Two Discrete
Levels}\label{gemini-3.1-pro-scores-at-only-two-discrete-levels}

Gemini 3.1 Pro ran five times, with total scores taking only two values:
three runs at 6 points and two runs at 19 points. Concurrently, in all
five runs, the value entered in its confidence field was 100 in all five
runs. Shifting scores alongside constant self-assessed confidence is a
behavioral characteristic worth recording in this batch of runs; this is
insufficient to determine how the model interprets its own confidence.

\subsubsection{5.1.4 Model Diagnostic
Profiles}\label{model-diagnostic-profiles}

Synthesizing the above dimensions, the six models in EP01 exhibited
distinguishable diagnostic profiles:

\begin{itemize}
\tightlist
\item
  \textbf{GPT-5.6 Sol}: Relatively high key sentence hits, poorer
  participant judgment, large score volatility.
\item
  \textbf{DeepSeek-V4-Flash}: Lower mean key sentence hits, highest mean
  correct participants, zero incorrect selections across all five runs.
\item
  \textbf{Gemini 3.1 Pro}: Total scores across five runs took only two
  values (6 and 19), with self-assessed confidence uniformly at 100.
\item
  \textbf{GLM-5.2}: Mean key sentence hits of 3.0, mean correct
  participants of 1.4 (tied with Gemini), mean incorrect selections of
  0.2.
\item
  \textbf{Claude Opus 4.6} and \textbf{Kimi K3}: Intermediate clue hits
  and participant selections, moderate volatility.
\end{itemize}

It must be emphasized that these profiles describe behavioral
characteristics of the models in this specific pilot experiment, and do
not constitute comprehensive evaluations of model reasoning ability. The
same model may exhibit different behavioral patterns under different
narrative structures and domain settings.

\subsection{5.2 Episode 02}\label{episode-02}

EP02 expanded the evaluation scale: 8 models × 5 runs = 40 formal runs,
of which 37 were content-scorable (3 runs were excluded due to invalid
selection content, see §4.7 for details). The full text contains 134
sentences, with 20 author-designated key clue sentences (ranging from 1
to 4 points each, maximum 41 points), 9 candidate characters, and a
theoretical maximum gameplay-equivalent score of 47 points. The case
structure of EP02 is more complex than EP01: the correct participant
chain involves multi-layered divisions of labor---planning, execution,
assistance, and cover-up---with necessary evidence dispersed across
low-salience, cross-paragraph, multi-functional information.

The table below reports each model's mean gameplay-equivalent total
score and subscale means.

{\def\LTcaptype{none} % do not increment counter
\begin{longtable}[]{@{}lrrr@{}}
\toprule\noalign{}
Model & Total Score Mean & Clue Score Mean & Participant Score Mean \\
\midrule\noalign{}
\endhead
\bottomrule\noalign{}
\endlastfoot
GPT-5.6 Sol & 6.80 & 14.80 & −8.00 \\
Claude Opus 4.6 & −1.00 & 12.20 & −13.20 \\
Gemini 3.1 Pro & 5.00 & 12.00 & −7.00 \\
GLM-5.3 & −6.40 & 10.00 & −16.40 \\
Kimi K3 & −8.00 & 11.50 & −19.50 \\
DeepSeek-V4-Pro & −8.60 & 9.00 & −17.60 \\
Hy4 preview & −1.25 & 14.75 & −16.00 \\
MiniMax-M3 & −9.00 & 7.00 & −16.00 \\
\end{longtable}
}

Note: Gemini 3.1 Pro, Kimi K3, and Hy4 preview have means calculated
across 4 content-scorable runs each, while remaining models have means
calculated across 5 runs; unparseable/non-scorable runs are not included
in means, nor are they scored as zero. Total score = clue score +
participant score; participant score includes rewards for correct
selections and deductions for incorrect selections, and may be negative.
Maximum scores for total score, clue score, and participant score are
47, 41, and 6 points, respectively. Raw scores between the two episodes
are not directly compared.

EP02 likewise exhibited marked process dissociation. The most
representative cases include: Hy4 preview's response03 hit 9 key clue
sentences (clue score 18), yet selected zero correct case participants;
Kimi K3's response03 hit a cohort-high 10 key clue sentences (clue score
17), but selected zero correct participants while misselecting 5
characters, yielding a participant score of −20 and a total score of −3.
In contrast, GPT-5.6 Sol's response01 hit only 5 key clue sentences
(clue score 12), yet was the sole run to select correct participants
without containing any incorrect characters (selecting Brother Lei and
Geng Baishun while missing Geng Baili), achieving the episode-high score
of 16/47 points. Identifying many clues does not equate to selecting
correct participants.

\subsubsection{5.2.1 Common Attractor: Convergent Errors Across
Models}\label{common-attractor-convergent-errors-across-models}

The most striking phenomenon in EP02 is the emergence of a
\textbf{common attractor}. Across 37 content-scorable runs:

\begin{itemize}
\tightlist
\item
  31/37 runs (83.8\%) included the trio of ``Father -- Jia Changgeng --
  Director Huang'' in participant selections;
\item
  16/37 runs (43.2\%) selected \textbf{strictly} this trio, neither more
  nor fewer;
\item
  18/37 runs (48.6\%) had this trio at the core, adding or omitting at
  most one character.
\end{itemize}

This convergence spanned all eight models and was not idiosyncratic
behavior of any single model. It should be clarified that none of these
three individuals were correct participants in EP02 (EP02's correct
participants were Brother Lei, Geng Baishun, and Geng Baili). Among
them, Jia Changgeng assumed a high-salience suspect role in the story
(substantial descriptions pointed to his suspicious behavior, −8
deduction), Director Huang held superficial connections to the case but
functioned in fact as a distractor (−6 deduction), and Father was a
correct participant in EP01 but not a participant in EP02 (−2
deduction).

The existence of a common attractor indicates that high-salience signals
in the story (such as multiple depictions of Jia Changgeng's suspicious
actions) can pull participant selections toward the same erroneous
direction across models and across runs. The significance of this
phenomenon lies in providing an empirical metric to test the
effectiveness of narrative design: if distractor structures preset by
professional authors can generate consistent misdirection effects across
disparate models, this in itself constitutes evidence of successful
not-fully-cooperative informational design.

\subsubsection{5.2.2 Evidence Asymmetry: S112
vs.~S113}\label{evidence-asymmetry-s112-vs.-s113}

EP02 provides a representative case of asymmetric processing between
positive and negative evidence. S112 and S113 are two adjacent sentences
in the story with diametrically opposed functions:

\begin{itemize}
\tightlist
\item
  \textbf{S113} (suspicion-oriented evidence): selected in 25/37 runs
  (67.6\%)
\item
  \textbf{S112} (exoneration-oriented evidence / alibi evidence):
  selected in 0/37 runs (0\%)
\end{itemize}

The two sentences are immediately adjacent in the text. The 25:0
selection frequency disparity demonstrates that a marked asymmetry
exists in model evidence selection between suspicion-oriented
information and exoneration-oriented information. It should be noted
that physical adjacency does not imply equal processing weights; S113
itself encompasses multiple pieces of information, such as the private
discussion act and the disappearance timeline, and specific rationales
for selecting it may vary across runs. What is reported here is the
disparity in selection frequencies, not a unified functional attribution
for every selection.

\subsubsection{5.2.3 S081 and Local Evidence-Selection
Covariation}\label{s081-and-local-evidence-selection-covariation}

S081 describes Geng Baishun's crutch anomaly. Across 37 content-scorable
runs, it was selected 6 times: GPT-5.6 Sol's response01, 03, 04, 05,
Claude Opus 4.6's response05, and Gemini 3.1 Pro's response02. All 6
runs simultaneously selected Brother Lei and Geng Baishun; Gemini's run
also selected Geng Baili, but additionally misselected Jia Changgeng.
Thus, what is observed here is a local co-occurrence between S081 and
Geng Baishun's selection, rather than six completely correct responses.
Because a single response submits sentence and participant selections
simultaneously, the temporal order in which the two selections arose, or
any causal relation between them cannot be determined from this alone.

\subsubsection{5.2.4 The S124 Phenomenon: Boundaries Between Puzzle
Premise and Scored
Clues}\label{the-s124-phenomenon-boundaries-between-puzzle-premise-and-scored-clues}

S124 describes Qin Guanggui being strangled from behind and hung from a
tree, along with the crime scene having only footprints of the deceased
leading toward the tree, with other footprints made later by those who
discovered the body. This sentence constitutes the core puzzle premise
of the case and provides constraints for inferring the modus operandi.
The author positioned it as a foundational fact requiring explanation,
and did not include it in the set of key clue sentences rewarding
specific inferential hubs.

All 37/37 content-scorable runs selected S124, earning no points for
this sentence under frozen rules. This unanimous selection shows that a
stable divergence exists between the evidentiary set submitted by models
and the author's scoring set: models universally selected the core
puzzle premise, whereas scoring emphasized other key sentences that
unravel the puzzle. While S124 exhibited no selection rate variance in
this batch of runs, its unanimous selection retains diagnostic
value---it reminds us to distinguish between ``failing to hit \(K\)''
and ``selecting information lacking inferential utility.''

\subsubsection{5.2.5 Five 0/37 Key Clue
Sentences}\label{five-037-key-clue-sentences}

Among the 20 key clue sentences in EP02, five were never selected by any
model across all 37 runs: S025, S074, S077, S106, and S112. However, the
nature of these omissions differs:

\begin{itemize}
\tightlist
\item
  \textbf{S112} was discussed in 5.2.2; its omission relates to the
  systematic neglect of exoneration-oriented evidence.
\item
  Neighboring sentences to \textbf{S025, S074, and S077} were selected
  in some runs; missing a specified sentence does not necessarily mean
  missing the entire segment of information. For example, MiniMax-M3's
  response02 did not select S025, but selected the adjacent S026, using
  the past event of Jia Changgeng deceiving the Geng family on behalf of
  the Huang family to infer Jia's relationship with the Huangs; that
  explanation did not explicitly utilize the old grievance between the
  Huang and Geng families stated in S025, so S026 cannot be treated as a
  fully equivalent substitute.
\item
  Although no model selected \textbf{S106}, one cannot judge solely from
  sentence selection that the role of fear was entirely absent from
  explanations. GPT-5.6 Sol's response01 selected S093 and S129,
  discussing terrifying paper figures and continuous intimidation;
  whether other runs utilized proximate information still requires
  inspecting explanations run by run.
\end{itemize}

Therefore, the raw 0/37 figure must be interpreted in conjunction with
neighboring sentence selections, and cannot be simply equated with
``complete omission.''

\subsubsection{5.2.6 Predominance of Negative Scores and Score
Distribution}\label{predominance-of-negative-scores-and-score-distribution}

Overall scores in EP02 were skewed low: 28/37 runs (75.7\%) yielded
negative gameplay-equivalent scores. The highest individual score was
GPT-5.6 Sol's response01, obtaining 16/47 points. The reason negative
scores predominated is that the correct participant chain in EP02
involved multi-layered division of labor, requiring integration of
evidence dispersed across low-salience positions; meanwhile, most runs
were drawn by the common attractor toward incorrect characters (such as
Jia Changgeng), whose penalty deductions exceeded the points gained from
hitting key clue sentences.

\subsubsection{5.2.7 Selecting Correct Characters and Eliminating
Incorrect Characters Are
Distinct}\label{selecting-correct-characters-and-eliminating-incorrect-characters-are-distinct}

At the participant selection level, EP02 further demonstrated the
dissociation between ``selecting correct characters'' and ``eliminating
incorrect characters'':

\begin{itemize}
\tightlist
\item
  15/37 runs (40.5\%) identified at least 1 correct case participant;
\item
  Yet among these 15 runs, 13 (86.7\%) still concurrently included the
  distractor character Jia Changgeng.
\end{itemize}

This indicates that \textbf{selecting correct characters} and
\textbf{eliminating incorrect characters} are two stages that can
manifest separately. Models can make headway in the correct
direction---indeed identifying some correct participants---while
remaining unable to resist the pull of high-salience distractor signals,
bundling Jia Changgeng into their selections as well. This aligns with
findings in EP01: reasoning tasks under the WORTH framework comprise
multiple sub-steps, and performing well on one step does not guarantee
performing well on another.

\subsubsection{5.2.8 EP02 Model Profiles}\label{ep02-model-profiles}

The diagnostic profiles of models in EP02 partially continued trends
from EP01, while displaying new bifurcations:

\begin{itemize}
\tightlist
\item
  \textbf{GPT-5.6 Sol}: Produced the highest-scoring run across the
  episode (16/47), but also produced a negative run (−3), continuing its
  high-volatility characteristic from EP01.
\item
  \textbf{DeepSeek-V4-Pro}: Total scores across five runs ranged from
  −12 to −6, all negative; averaged 4.2 incorrect selections, only
  slightly lower than Kimi's 5.0, with participant selection not
  continuing the conservative and precise profile of DeepSeek-V4-Flash
  in EP01.
\item
  Newly added \textbf{Hy4 preview} and \textbf{MiniMax-M3} provided
  additional observation points in EP02, but cross-episode comparisons
  are unwarranted given the pilot scale.
\end{itemize}

\subsection{5.3 Structural Patterns Shared Across Both
Episodes}\label{structural-patterns-shared-across-both-episodes}

Raw scores between EP01 and EP02 cannot be directly compared---the two
episodes differ in text length (120 vs.~134 sentences), number and point
structure of key clue sentences (12 × 3 points vs.~20 × 1--4 points),
case complexity and candidate character count (6 vs.~9 characters), and
participating models do not completely overlap (6 vs.~8 models).
Consequently, the following cross-episode comparison focuses on
\textbf{whether structural patterns appear in both episodes}, rather
than direct numerical comparisons.

\subsubsection{5.3.1 Cross-Episode Consistency of Process
Dissociation}\label{cross-episode-consistency-of-process-dissociation}

Both episodes feature cases where clue hits and participant selections
are desynchronized: in EP01, GPT-5.6 Sol's mean key sentence hits
exceeded DeepSeek-V4-Flash, yet its mean correct characters were lower;
in EP02, Kimi K3's response03 hit 10 key sentences, yet selected zero
correct characters. ``Process dissociation'' here is a description of
output discrepancies; cases across two episodes are insufficient to rule
out the influence of story or prompt configurations.

\subsubsection{5.3.2 Cross-Episode Consistency of Positive-Negative
Evidence
Asymmetry}\label{cross-episode-consistency-of-positive-negative-evidence-asymmetry}

Both episodes exhibited asymmetric model processing of
``suspicion-oriented evidence'' versus ``exoneration-oriented
evidence'':

\begin{itemize}
\tightlist
\item
  In EP01, S064 and S072, used by the author to eliminate Qin Guanggui
  and Wang Yonghong as direct perpetrators, were both 0/30; S094,
  involving Director Huang arranging cupping, was selected in 22/30
  runs.
\item
  In EP02, the 0:25 contrast between S112 and S113 provided more precise
  quantitative evidence.
\end{itemize}

The joint emergence of this asymmetry across both episodes suggests that
it may represent a structural tendency of current large language models
when processing mystery narratives. However, it must be noted: this
observation is grounded in a specific domain (mystery narratives) and
specific designs (informational manipulation in professional
narratives), and generalizing it to other domains requires further
empirical verification.

\subsubsection{5.3.3 Systematic Omission of ``I'' and Unanimous
Selection of
S124}\label{systematic-omission-of-i-and-unanimous-selection-of-s124}

Correct character R04 in EP01 was selected by no model across 30 runs;
S124, which was outside the author's key clue sentence set in EP02, was
selected by all 37 content-scorable runs. The two extreme selections
point respectively to construct boundary issues regarding participant
definitions and the scope of scored clues. Raw explanations and
potential causes are discussed in §6.5 and §6.3, respectively.

\section{6. Discussion}\label{discussion}

Section 5 reported the observable behavioral structures in the
two-episode pilot experiment. This section discusses the implications of
these results: why they provide diagnostic information not easily
revealed by traditional reasoning evaluations, what empirical feedback
they offer for WORTH's design assumptions, and which explanations still
require subsequent experiments to disentangle.

\subsection{6.1 Process Dissociation: Phenomena Manifested Under
Autonomous Evidence
Discovery}\label{process-dissociation-phenomena-manifested-under-autonomous-evidence-discovery}

Itemized outputs across both episodes show that key sentence hits and
participant judgments can be desynchronized. The EP01 model mean
contrasts and EP02 run-level cases in Section 5 support reporting these
two classes of results separately; they do not isolate internal
reasoning steps within models.

Evaluations that report only final answer scores struggle to present
performance in evidence discovery and participant judgment separately.
When evidence is provided by the test designer, the front end of the
reasoning chain (judging what is worth becoming evidence) is substituted
by the test designer, allowing evaluation to see only the latter half
from known premises to conclusions. WORTH's autonomous evidence
discovery condition renders this front-end stage an observable output as
well. For instance, in EP01, GPT-5.6 Sol hit more key sentences yet
correctly identified fewer characters, while DeepSeek-V4-Flash displayed
the inverse combination. Looking solely at total scores would obscure
this disparity; recording both outputs separately brings it to light.
Measuring the two stages separately helps differentiate model
performance variations across distinct sub-processes, rather than
observing only a conflated signal.

Of course, process dissociation has currently been observed only in two
mystery narratives, and a scale of 70 runs is insufficient to establish
stable quantitative estimates. Whether it exists similarly across other
types of narratives and reasoning tasks requires more data to answer.
Yet even if treated only as a recurring behavioral pattern, it
demonstrates that in measuring reasoning ability, ``what is found'' and
``what judgment is made using what was found'' deserve separate
investigation.

\subsection{6.2 Common Attractor: Empirical Metric of Narrative Design
Effectiveness}\label{common-attractor-empirical-metric-of-narrative-design-effectiveness}

In EP02, 31/37 runs included the ``Father -- Jia Changgeng -- Director
Huang'' trio in participant selections, spanning all eight models and
displaying a concentrated erroneous selection pattern. Jia Changgeng was
designed by the author as a high-salience suspect---substantial
descriptions pointed to his suspicious conduct, and multiple clues
superficially supported the judgment that he participated in the crime.
Director Huang held superficial connections to the case but functioned
in fact as a distractor. That 83.8\% of runs contained the same set of
incorrect characters aligns with the author's intended
suspicion-steering direction, providing a concrete entry point for
analyzing how narrative interference affects participant judgment.

This phenomenon provides direct empirical support for WORTH's second
conditional shift---not-fully-cooperative informational relationships.
Section 3 argued why professionally authored mystery narratives serve as
appropriate experimental vehicles: authors achieve two tasks
concurrently---providing evidence in principle sufficient to deduce the
truth, and delaying the revelation of truth through narrative
techniques. The common attractor links the author's suspicion-steering
design with actual model selections, representing an empirical
phenomenon worth analyzing in not-fully-cooperative information
environments. Some responses also analyzed both Uncle Six's case and
subsequent death events concurrently (see §4.9), necessitating
interpreting participant selections in light of their actual
explanations. Subsequent control conditions can further test the impact
of narrative obfuscation on this convergence.

However, alternative explanations must be preserved. The emergence of a
common attractor can be attributed to at least three mechanisms, which
current data cannot disentangle:

The first possibility is insufficient epistemic vigilance in models.
Confronted with a superficially coherent explanatory framework, models
do not inquire ``whether this explanation missed something'' or
``whether counter-evidence was ignored''; once a framework explaining
most phenomena is assembled, they halt there.

The second possibility is that salience control in the story design was
overly potent. Depictions of Jia Changgeng's suspicion overwhelmed
counter-evidentiary information in volume and intensity, making it
difficult even for subjects with critical capacity to resist. In other
words, the problem may lie not in the models, but in the test's
distractor design being ``too effective.''

The third possibility is a joint effect---insufficient epistemic
vigilance in models compounded by potent salience control jointly
produced highly convergent outcomes.

The evidence-cueing condition proposed in §3.4 can observe whether
participant attribution shifts after key sentences are marked; the
evidence-modification condition can observe how answers shift alongside
alterations to key facts. Neither has been implemented yet, nor can any
single control attribute the common attractor solely to story-text salience
or model judgment styles. To specifically test the salience of suspicion
descriptions, rewritten conditions that hold case facts constant would
also need to be established.

\subsection{6.3 Selection Boundaries Between Puzzle Premise and
Evidence}\label{selection-boundaries-between-puzzle-premise-and-evidence}

The unanimous selection of S124 illustrates a boundary between author
scoring design and model evidence selection. The author positioned it as
the core puzzle premise, placing scoring emphasis on reasoning hubs that
help crack the puzzle; models, in contrast, universally incorporated
this sentence describing the death method and footprint constraints into
evidence. These two orientations do not entirely coincide, yet the
puzzle premise itself can constrain explanations; one cannot conclude
that models conflated informational functions solely because S124 was
not in \(K\).

Raw explanations provide concrete examples: GPT-5.6 Sol's response01
used S124 to rule out suicide and treated concealing footprints of
companions as the core of solving the puzzle. The prompt also instructed
selecting sentences highly relevant to motives, methods, or participant
identities, providing clear textual justification for selecting S124.
The 37/37 consistency indicates that this selection is robust, yet is
insufficient by itself to determine whether models distinguished between
the phenomenon to be explained and the critical clues that crack the
puzzle.

This phenomenon does not warrant modifying the current frozen scoring.
\(K\) is used to measure hits on preset reasoning hubs, while model
selections and explanations provide richer diagnostic information. Only
by combining both can one observe what models selected, how they used
it, and where divergences lie relative to scoring objectives.

\subsection{6.4 Evidence Asymmetry and Epistemic
Vigilance}\label{evidence-asymmetry-and-epistemic-vigilance}

S112 and S113 are immediately adjacent in the story. S113 points toward
suspicion and was selected in 25/37 runs; S112 provides
exoneration-oriented evidence (alibi evidence) and was selected in 0/37
runs. The two sentences are physically contiguous in the text. The 25:0
frequency disparity reveals that suspicion-oriented information enters
model evidence selection far more readily than exoneration-oriented
information. Concurrently, it should be noted that S112's phrasing
includes the qualifier ``most of the time,'' rendering its exclusionary
force weaker than an absolute alibi, which may partially account for its
low selection rate; nonetheless, the 0/37 selection count remains a
striking asymmetry.

This asymmetry is manifested across both episodes. In EP01,
high-salience information related to conspiracy motives was broadly
selected, while low-salience clues pointing to exoneration were rarely
noted. In EP02, S112/S113 provided a concrete case of adjacent positions
with starkly divergent selection frequencies. Cross-episode consistency
suggests that asymmetric processing of positive and negative evidence
may reflect a structural tendency of current models when handling mystery
narratives, rather than an incidental artifact of one episode's story
structure.

The epistemic vigilance framework proposed by Sperber et al.~provides an
interpretive lens for this observation. Story texts in WORTH
simultaneously contain information facilitating reasoning and
information delaying reasoning; when models incorporate suspicion
information while omitting clues used by authors to exonerate suspects,
one possibility is that they accept superficially coherent explanations
without adequately scrutinizing counter-evidence. This makes epistemic
vigilance a direction worthy of further investigation.

However, boundaries of this observation must be made explicit. Current
data originate from the mystery-narrative domain, where the pair of
directions ``suspicion'' and ``exoneration'' carries domain specificity.
Generalization to other domains---such as counterexample identification
in scientific literature, exculpatory evidence assessment in legal
argumentation, or risk signals in business decision-making---requires
independent verification in those domains. Currently, one can only state
that this asymmetry exists in mystery narratives, without confirming
whether it reflects a general model trait or a domain-specific
behavioral pattern.

\subsection{6.5 Boundaries of Narrative
Perspective}\label{boundaries-of-narrative-perspective}

In EP01, the character appearing in the first person as ``I'' (R04) was
a correct participant designated by the author, yet across 30/30
runs---spanning all six models---was not selected as a participant a
single time. In contrast, S124 was selected by all content-scorable runs
in EP02, despite not being in \(K\). These two extreme selections touch
respectively upon participant definitions and the boundary between core
puzzle premises and scored clues.

Narrative perspective may influence model participant judgment, but raw
responses provide more concrete clues: some runs actually noticed ``my''
causal role in the case---for instance, Kimi response04 termed ``I'' an
``innocent tool being exploited,'' and DeepSeek response05 and Gemini
response05 mentioned the narrator's cupping act in explanations---yet
still did not select R04 as a participant. The test prompt listed R04
independently as a candidate character and provided a
\texttt{selected\_role\_ids} field; selecting characters did not require
locating a specific sentence as a carrier, making it unlikely that
omission stemmed from selection mechanism constraints. A more probable
explanation is that models adopted a definition of ``participant''
bounded by awareness or subjective intent, whereas the author's correct
character set encompassed unwitting causal facilitation. Another
possibility is that the presentation of the candidate character list in
response instructions influenced attention allocation. Current data
cannot fully disentangle these hypotheses; systematic variations in
narrative perspective in subsequent experiments (such as rewriting the
same story in third person) can provide further evidence.

\subsection{6.6 Design Justification and Pilot
Feedback}\label{design-justification-and-pilot-feedback}

Synthesizing results from Section 5 allows revisiting the design
justification of the key clue sentence set and clarifying considerations
regarding scoring disclosure.

\textbf{Validity of the Key Clue Sentence Set.} \(K\) does not consist
of ``sentences that seem important'' annotated post-hoc by researchers
from the story; rather, they are reasoning hubs planted in advance
during the design phase by creators. The studio's creative workflow
specifies character trajectories and event logic first, after which
narrative techniques are used to embed and obscure them---\(K\) emerges
during the event design phase, reflecting the creative intent of ``which
informational nodes must be traversed to infer from A to B.'' Narrative
wrapping alters how conspicuous these nodes appear in the text, without
changing their logical function. This means the epistemological status
of \(K\) is closer to grading rubrics in examination design: test
creators need not prove that no alternative solution path exists in the
world; what requires demonstration is that this scoring rubric is
internally consistent and capable of differentiating responses of
varying quality. Pilot results provide concrete cases for examining
these designs: five zero-hit key clue sentences in EP02 precisely
covered inferential steps required to rule out false hypotheses and
complete the correct participant chain, with runs missing them
systematically converging toward erroneous character
combinations---these sentences occupied critical positions in ruling out
false hypotheses within the author's designed reasoning chain. While
unanimous zero hits mean their empirical discriminatory power cannot be
validated within this sample batch, their design-level logical function
aligns with observed error directions. S124 presents another class of
information: it was selected by all runs; appending a fixed point value
to this existing batch of selections would add points equally across all
runs. This fact describes its discrimination within this sample batch,
and does not independently dictate whether it should enter \(K\); the
author's basis for excluding it remains the aforementioned scoring
objective (see §6.3).

\textbf{Boundaries of Scoring Information Disclosure.} As noted in §4.4,
the prompt informs subjects of reward and penalty directions for clue
and participant selections, but conceals specific weights. This design
aims to reduce gaming strategies around point values; no disclosure
control experiment has yet tested its actual effect. Instructions for
EP02 already provided an operational definition of ``homicide
participant'' (see §4.3).

\subsection{6.7 Human Data}\label{human-data}

This version does not report human scores or human-model comparisons.
Once verifiable data are obtained following the criteria in §4.8,
subsequent work can examine whether phenomena such as shared participant
selections and S124 selection rates also appear in player responses,
aiding in distinguishing item effects from model-specific behaviors;
such judgments cannot be made at present.

\section{7. Scope of Research and Subsequent
Verification}\label{scope-of-research-and-subsequent-verification}

\subsection{7.1 Scope of Applicability of
Results}\label{scope-of-applicability-of-results}

This pilot experiment comprises two story episodes and 70 formal model
runs, with each model run 5 times per episode. On this basis, this paper
reports scores, selection patterns, and specific errors of models under
current task and operational conditions; the main text cites only
selected run-by-run cases. Model means serve to summarize performance in
this execution, and are not used to establish general capability
rankings. Subsequent additions of stories and run counts can test the
replicability of observed patterns across more tasks.

\subsection{7.2 Content Covered by Current
Scoring}\label{content-covered-by-current-scoring}

Gameplay-equivalent total scores consist of clue scores and participant
scores, quantifying outcomes of responding under frozen rules. Clue
scores take \(K\) as reference; the phenomenon of neighboring sentence
selection reported in Section 5.2.5 indicates that missing designated
sentences does not necessarily equal missing corresponding information.
Natural language explanations have been preserved and used for case
analysis, but have not been scored independently, nor have inter-rater
reliability tests across independent coders been conducted. Future work
could conduct functional coding on neighboring selections and
explanation content to disentangle sentence selection variance,
information omission, and reasoning errors.

\subsection{7.3 Subsequent Verification}\label{subsequent-verification}

Section 3.4 proposed two classes of controls not yet implemented: first,
retaining the complete story while highlighting author-designated key
clue sentences, comparing shifts in participant attribution; second,
modifying key evidence while maintaining story coherence and altering
answers accordingly, observing whether model judgments shift alongside
evidence changes. These test the effect of evidence cueing and model
responsiveness to evidence shifts, respectively.

Player records exist on game platforms for both episodes, but export and
cleaning remain incomplete; this version does not report human scores or
human-model comparisons. After verifiable data are obtained in the
future, same-task reference analyses will be conducted. For model
testing conducted after materials become public, material release dates
and model versions should be recorded, and combined with new stories and
evidence-modification conditions to examine result replicability.

\section{8. Conclusion}\label{conclusion}

This paper presents WORTH, introducing autonomous evidence discovery and
not-fully-cooperative informational relationships into large language
model reasoning evaluation. Through professionally crafted natural
narratives, WORTH requires models to autonomously select key clues,
identify case participants, and explain their reasoning process from
complete stories, separately recording performance in evidence selection
and participant attribution.

Across two episodes, 70 formal runs were conducted, of which 67 runs
were content-scorable. Pilot results demonstrate the diagnostic value of
this evaluation framework: hitting more key clues does not guarantee
more accurate participant judgments; different models converge on
selecting identical distractor characters; and information pointing
toward suspicion versus exoneration exhibits marked selection asymmetry.
Total scores and their subscales present scoring outcomes of these
selections, while specific sentences, characters, and explanations
exhibit traceable discrepancies in erroneous judgments.

These results indicate that observing evidence selection and final
judgment separately can reveal reasoning behavioral differences
difficult to convey through a single total score. Subsequent work will
further examine these patterns through additional stories,
evidence-cueing, and evidence-modification conditions. What WORTH
emphasizes is that reasoning evaluation must examine how models identify
evidence from information, and how they transform evidence into
judgment.

\section{Acknowledgments}\label{acknowledgments}

We especially thank Shuwen Yang, Zhiyuan Zhu, and Weimin Zong for their
help with data organization, statistical testing, providing model API
access, and assisting with model testing.

\section{Appendix A. Compact Experimental
Specification}\label{appendix-a.-compact-experimental-specification}

\subsection{A.1 Task and Response
Protocol}\label{a.1-task-and-response-protocol}

Each run received, in a single turn, the complete Chinese story with
neutral sentence identifiers, the candidate-character list, and the
response instructions. The model selected up to the episode-specific
clue quota, selected zero or more candidate characters without being
told the number of correct participants, and returned seven fields:
\texttt{episode\_id}, \texttt{clue\_sentence\_ids},
\texttt{selected\_role\_ids}, \texttt{clue\_interpretations},
\texttt{evidence\_combination}, \texttt{final\_conclusion}, and
\texttt{confidence}. Clue interpretations were required to correspond
one-to-one with the selected sentence identifiers. Runs used fresh
conversations, with search, external tools or connectors, memory, and
personalization disabled. Natural-language explanations were retained
for diagnosis but were not assigned an independent score.

\subsection{A.2 Episode-Level Scoring
Summary}\label{a.2-episode-level-scoring-summary}

{\def\LTcaptype{none} % do not increment counter
\begin{longtable}[]{@{}
  >{\raggedright\arraybackslash}p{(\linewidth - 8\tabcolsep) * \real{0.1875}}
  >{\raggedright\arraybackslash}p{(\linewidth - 8\tabcolsep) * \real{0.1875}}
  >{\raggedright\arraybackslash}p{(\linewidth - 8\tabcolsep) * \real{0.1875}}
  >{\raggedright\arraybackslash}p{(\linewidth - 8\tabcolsep) * \real{0.1875}}
  >{\raggedleft\arraybackslash}p{(\linewidth - 8\tabcolsep) * \real{0.2500}}@{}}
\toprule\noalign{}
\begin{minipage}[b]{\linewidth}\raggedright
Episode
\end{minipage} & \begin{minipage}[b]{\linewidth}\raggedright
Story and selection space
\end{minipage} & \begin{minipage}[b]{\linewidth}\raggedright
Clue scoring
\end{minipage} & \begin{minipage}[b]{\linewidth}\raggedright
Participant scoring
\end{minipage} & \begin{minipage}[b]{\linewidth}\raggedleft
Theoretical total range
\end{minipage} \\
\midrule\noalign{}
\endhead
\bottomrule\noalign{}
\endlastfoot
EP01 & 120 sentences; at most 12 clue selections; 6 candidates, 3
correct participants & 12 key sentences, 3 points each (36 total) &
Correct participants: +7, +5, +5; each incorrect participant: −5 & −15
to 53 \\
EP02 & 134 sentences; at most 20 clue selections; 9 candidates, 3
correct participants & 20 key sentences, weighted 1--4 points (41 total)
& Each correct participant: +2; incorrect participants: −2 to −8 by
character & −22 to 47 \\
\end{longtable}
}

In both episodes, omitting a key sentence and selecting a sentence
outside the author-designated set \(K\) received zero clue points rather
than a penalty. The two episodes use different stories and payoff
structures; their raw scores are therefore reported separately and are
not summed or treated as a common difficulty scale.

\subsection{A.3 Model and Run
Configurations}\label{a.3-model-and-run-configurations}

Each row below represents five formal runs. Product and tier labels
reproduce the settings recorded at execution time; they should not be
interpreted as standardized compute levels across vendors.

{\def\LTcaptype{none} % do not increment counter
\begin{longtable}[]{@{}
  >{\raggedright\arraybackslash}p{(\linewidth - 8\tabcolsep) * \real{0.2000}}
  >{\raggedright\arraybackslash}p{(\linewidth - 8\tabcolsep) * \real{0.2000}}
  >{\raggedright\arraybackslash}p{(\linewidth - 8\tabcolsep) * \real{0.2000}}
  >{\raggedright\arraybackslash}p{(\linewidth - 8\tabcolsep) * \real{0.2000}}
  >{\raggedright\arraybackslash}p{(\linewidth - 8\tabcolsep) * \real{0.2000}}@{}}
\toprule\noalign{}
\begin{minipage}[b]{\linewidth}\raggedright
Episode
\end{minipage} & \begin{minipage}[b]{\linewidth}\raggedright
Model
\end{minipage} & \begin{minipage}[b]{\linewidth}\raggedright
Access interface
\end{minipage} & \begin{minipage}[b]{\linewidth}\raggedright
Recorded tier or mode
\end{minipage} & \begin{minipage}[b]{\linewidth}\raggedright
Test window
\end{minipage} \\
\midrule\noalign{}
\endhead
\bottomrule\noalign{}
\endlastfoot
EP01 & GPT-5.6 Sol & ChatGPT web & High & 2026-08-14 \\
EP01 & Claude Opus 4.6 & Antigravity & Thinking & 2026-08-14 \\
EP01 & GLM-5.2 & Zhipu web & High (Deep Thinking) & 2026-08-14 \\
EP01 & DeepSeek-V4-Flash & TraeWork app & V4-Flash & 2026-08-14 \\
EP01 & Kimi K3 & Kimi website & Advanced & 2026-08-14 \\
EP01 & Gemini 3.1 Pro & Antigravity IDE & High & 2026-08-14 \\
EP02 & GPT-5.6 Sol & ChatGPT web & High & 2026-08-30 \\
EP02 & Claude Opus 4.6 & Claude.ai & High + Extended & 2026-08-30 to
2026-08-31 \\
EP02 & Gemini 3.1 Pro & Antigravity IDE & High & 2026-08-30 \\
EP02 & GLM-5.3 & Zhipu website & Deep & 2026-08-30 \\
EP02 & Kimi K3 & WorkBuddy for macOS & High & 2026-08-30 \\
EP02 & DeepSeek-V4-Pro & TraeWork app & Not separately recorded &
2026-08-30 \\
EP02 & Hy4 preview & WorkBuddy for macOS & High & 2026-08-30 \\
EP02 & MiniMax-M3 & WorkBuddy for macOS & n/a & 2026-08-30 \\
\end{longtable}
}

For EP01 GLM-5.2, the configuration and execution screenshots showed
``Highest'' and ``High,'' respectively; the table follows the execution
record and reports High, with Deep Thinking enabled. The EP02 DeepSeek
tier was not separately recorded, and the MiniMax-M3 interface exposed
no applicable tier label.

\subsection{A.4 Run Validity and
Exclusions}\label{a.4-run-validity-and-exclusions}

{\def\LTcaptype{none} % do not increment counter
\begin{longtable}[]{@{}
  >{\raggedright\arraybackslash}p{(\linewidth - 12\tabcolsep) * \real{0.1111}}
  >{\raggedleft\arraybackslash}p{(\linewidth - 12\tabcolsep) * \real{0.1481}}
  >{\raggedleft\arraybackslash}p{(\linewidth - 12\tabcolsep) * \real{0.1481}}
  >{\raggedleft\arraybackslash}p{(\linewidth - 12\tabcolsep) * \real{0.1481}}
  >{\raggedleft\arraybackslash}p{(\linewidth - 12\tabcolsep) * \real{0.1481}}
  >{\raggedleft\arraybackslash}p{(\linewidth - 12\tabcolsep) * \real{0.1481}}
  >{\raggedleft\arraybackslash}p{(\linewidth - 12\tabcolsep) * \real{0.1481}}@{}}
\toprule\noalign{}
\begin{minipage}[b]{\linewidth}\raggedright
Episode
\end{minipage} & \begin{minipage}[b]{\linewidth}\raggedleft
Formal runs
\end{minipage} & \begin{minipage}[b]{\linewidth}\raggedleft
Directly parseable JSON
\end{minipage} & \begin{minipage}[b]{\linewidth}\raggedleft
Embedded parseable JSON
\end{minipage} & \begin{minipage}[b]{\linewidth}\raggedleft
Invalid overall JSON
\end{minipage} & \begin{minipage}[b]{\linewidth}\raggedleft
Content-scorable runs
\end{minipage} & \begin{minipage}[b]{\linewidth}\raggedleft
Fully protocol-compliant runs
\end{minipage} \\
\midrule\noalign{}
\endhead
\bottomrule\noalign{}
\endlastfoot
EP01 & 30 & 22 & 7 & 1 & 30 & 22 \\
EP02 & 40 & 26 & 9 & 5 & 37 & 25 \\
Total & 70 & 48 & 16 & 6 & 67 & 47 \\
\end{longtable}
}

Parseability and content validity were evaluated separately. The single
EP01 response with invalid overall JSON still contained directly
readable selection arrays, so all 30 EP01 runs remained
content-scorable. In EP02, four of the five responses with invalid
overall JSON retained readable, valid selection arrays. Three EP02 runs
were excluded from content scoring:

\begin{itemize}
\tightlist
\item
  Gemini 3.1 Pro response05 selected the nonexistent sentence identifier
  \texttt{S207}.
\item
  Hy4 preview response05 selected the nonexistent sentence identifier
  \texttt{S146}.
\item
  Kimi K3 response02 used illegal expressions in the selection fields,
  so valid identifier arrays could not be recovered.
\end{itemize}

Excluded runs were not assigned a score of zero. One additional EP01
Codex run was retained only as an exploratory auxiliary control; it was
not part of the formal model set and is not included in the 70 formal
runs reported in this paper.

\bibliographystyle{plainnat}
\bibliography{references}

\end{document}